\documentclass{article}

\usepackage[T1]{fontenc}
\usepackage[accepted]{mlsys2025}

\makeatletter
\renewcommand{\Notice@String}{}
\makeatother

\usepackage{amsmath}
\usepackage{amssymb}
\usepackage{booktabs}
\usepackage{multirow}
\usepackage{graphicx}
\usepackage{subcaption}
\usepackage{xspace}
\usepackage{microtype}
\usepackage{enumitem}
\usepackage{url}
\usepackage{xcolor}
\usepackage{siunitx}
\usepackage{algorithmic}
\usepackage{svg}

\renewcommand{\algorithmiccomment}[1]{%
  \hfill$\triangleright$~\textit{#1}%
}

\newcommand{\method}{FastGRT\xspace}

\mlsystitlerunning{FastGRT: Accelerating Training of 3D Gaussian Ray Tracing}

\begin{document}

\twocolumn[
\mlsystitle{
FastGRT: Accelerating Training of 3D Gaussian Ray Tracing
}

\mlsyssetsymbol{equal}{*}


\begin{mlsysauthorlist}
\mlsysauthor{Meet Banthia}{nyu,equal}
\mlsysauthor{Xiaoteng Liu}{nyu,equal}
\mlsysauthor{Hexu Zhao}{nyu}
\mlsysauthor{Aurojit Panda}{nyu}
\mlsysauthor{Jinyang Li}{nyu}
\end{mlsysauthorlist}

\mlsysaffiliation{nyu}{New York University}

\mlsyscorrespondingauthor{Meet Banthia}{mjb9520@nyu.edu}

\mlsyskeywords{
3D Gaussian Ray Tracing,
Differentiable Rendering,
GPU Optimization,
Vulkan
}

\vskip 0.3in
]

\printAffiliationsAndNotice{\mlsysEqualContribution}

\begin{abstract}



Ray Tracing enables rendering capabilities that are difficult to achieve with rasterization-based approaches such as 3D Gaussian Splatting (3DGS). However,  the training cost of Gaussian-based ray tracing (GRT) is much higher than that of 3DGS. 

In this paper, we study how ray tracing using Gaussians executes on GPUs and show that a considerable fraction of the training cost arises from how the computation is organized. We develop {\method}, which decouples ray traversal from blending and reuses work done in the forward pass for the backward pass. {\method} further incorporates optimizations such as fast gradient accumulation and a hybrid BVH update strategy to tackle other major bottlenecks in the training pipeline.

We implement {\method} in Vulkan. To the best of our knowledge, it is the first Vulkan-based GRT implementation that supports both training and rendering. Compared to 3DGRT, \method improves training time by $1.9-4.97\times$, while preserving the original rendering formulation. Our code is open sourced
\footnote{\url{https://github.com/meetbanthia/fastgrt}}.
\end{abstract}

\section{Introduction}


Novel view synthesis (NVS) reconstructs and learns a 3D scene from a collection of observed images from certain viewpoints and render photorealistic images from previously unseen viewpoints. Recent advances in 3D Gaussian Splatting~\cite{kerbl20233dgs} (3DGS) have made this NVS highly efficient both in terms of speed and quality. 3DGS is a rasterization-based algorithm. Its efficiency comes from projecting each 3D Gaussian into screen space using geometric transformations, which maps extremely well to modern GPUs. Applications like autonomous cars require wide-angle cameras (e.g., fisheye cameras), whose lenses are prone to damage, which leads to distortion. Real cameras are often not ideal pinhole cameras; for such complex camera models, the projected 2D Gaussian no longer represents the underlying 3D Gaussian with minimal approximation, and this becomes a source of reconstruction error for complex models. Additionally, consider a camera that may observe an object only through its reflection in a mirror. Here, light from the object to the camera does not follow a straight path; instead, it is reflected by the mirror. This type of secondary light effect is difficult to represent with direct projections used by splatting.

\begin{figure}[t]
    \centering
    \includegraphics[width=0.75\columnwidth]{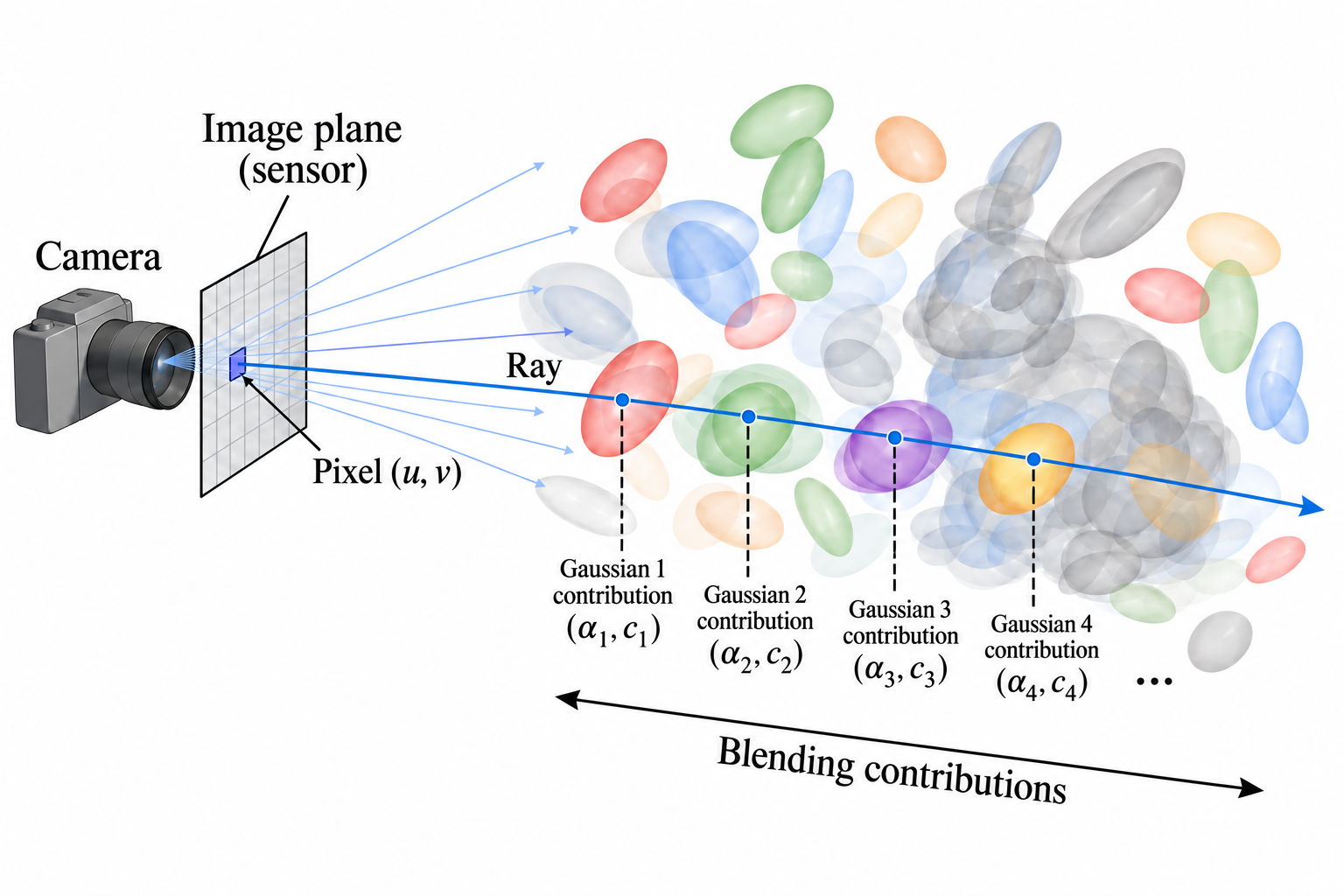}
    \caption{Ray tracing using 3D Gaussians. A ray is cast from each pixel and the hit Gaussians' contributions are blended together to produce the pixel color.}
    \label{fig:raytracing}
\end{figure}

Ray tracing solves the problem from the opposite direction (Figure~\ref{fig:raytracing}): instead of projecting every scene element onto the image plane, it launches a ray for each pixel and queries which scene elements intersect that ray. Arbitrary camera models can therefore be expressed simply by changing the ray origin and direction, while additional rays can be spawned at interactions to model reflection, refraction, and shadowing. 3D Gaussian Ray
Tracing (3DGRT)~\cite{moenneloccoz20243dgrt} brings this formulation to Gaussian scenes, replacing Gaussian projections with direct evaluation of Gaussians along rays.

This flexibility comes at the cost of rendering and training speed. Modern ray-tracing systems organize geometry using spatial acceleration structures like BVH (Bounding Volume Hierarchy) for efficiently finding ray--Gaussian hits. Dedicated hardware like NVIDIA RT cores, AMD ray-tracing accelerators, and Intel ray-tracing units helps accelerate BVH traversal. Despite the acceleration, computing intersection remains a major cost. 

Existing Gaussian ray tracing systems such as 3DGRT retrace 
the Gaussian hits during its backward pass, even though 
the forward pass has already traced each camera ray and discovered the Gaussians contributing to its pixel. 
However, reusing the forward-pass hits instead of re-tracing is not straightforward: the number of hits per ray is unknown in advance and varies widely across pixels and training iterations, and recording them inside the tracing kernel adds memory traffic that can slow the forward pass enough to cancel the savings.  
Moreover, many adjacent rays compute gradients of the same Gaussian, requiring large numbers of global atomic accumulation operations, resulting in large synchronization overhead. BVH maintenance can also be costly. A BVH represents the spatial layout of the Gaussians, but this layout continuously changes to better fit the 3D scene. Exising systems update BVH structure at each training iteration. So for large scenes containing millions of Gaussians, this maintenance becomes a considerable fraction of training time.



These costs stem from how training is organized around ray traversal rather than from traversal itself, so they can be reduced without changing what 3DGRT computes. Instead of retracing rays, our forward pass keeps the hits it already discovers, and splits tracing into short rounds so that recording them does not slow it down. Instead of accumulating every gradient directly to global memory, neighboring rays that update the same Gaussian first combine their contributions on-chip. And instead of rebuilding the BVH from scratch every iteration, we refit it in place whenever the set of Gaussians is unchanged. Each of these changes preserves the Gaussians blended for every pixel, so our system follows the rendering formulation of 3DGRT exactly while training substantially faster.

To study these mechanisms, we implement our method in Vulkan~\cite{vulkan14spec}. In contrast to prior OptiX-based implementations~\cite{10.1145/1778765.1778803}, Vulkan is cross-vendor and exposes lower-level control over hardware.

In summary, we make the following contributions:

\begin{itemize}[leftmargin=*]
    \item We eliminate retracing in the backward pass. With \emph{adaptive multi-round traversal}, the forward pass records every hit in a coalesced per-pixel archive without capping the number of rounds per ray, so the backward pass replays these hits instead of traversing the BVH again.

    \item We reduce the two other major training costs. A \emph{fast gradient accumulator} combines the gradients of rays that update the same Gaussian in on-chip memory before issuing global atomics, speeding up the backward pass by up to 10$\times$ at full resolution. A \emph{hybrid BVH update strategy} refits the BVH in most iterations and rebuilds it only when the Gaussian population changes or periodically thereafter.

    \item We implement these techniques in \method, the first open-source\footnote{GRTX~\cite{lee2026grtxefficientraytracing} also implements its method in Vulkan, but its code is not publicly available. Other Vulkan projects, such as vk\_gaussian\_splatting~\cite{vkgaussiansplatting}, support rendering only, without training.} Vulkan implementation of 3DGRT that supports both training and rendering. Across Mip-NeRF 360, Tanks and Temples, Deep Blending, and Zip-NeRF, \method trains 1.9--4.97$\times$ faster than 3DGRT while matching its rendering quality and performance.
\end{itemize}

\section{Background and Motivation}
\label{sec:background}

This section gives an overview of the standard 3D Gaussian ray tracing (3DGRT) algorithm~\cite{moenneloccoz20243dgrt} and its training pipeline, discusses its bottlenecks, and explains the optimization challenges.

\subsection{3D Gaussian Scene Primitive}
3D Gaussian rendering algorithms represent a scene with a large collection of anisotropic semi-transparent Gaussian primitives (aka ellipsoids)~\cite{kerbl20233dgs} distributed in the 3D world space. Each Gaussian $i$ is parameterized by a mean position $\boldsymbol{\mu}_i$, covariance $\boldsymbol{\Sigma}_i$, opacity $\alpha_i$, and view-dependent color represented by spherical harmonics (SH). 
In practice, each Gaussian is truncated to a finite ellipsoidal extent. The covariance is parameterized as
$\boldsymbol{\Sigma}_i=\mathbf{R}_i\mathbf{S}_i\mathbf{S}_i^T\mathbf{R}_i^T$,
where $\mathbf{R}_i$ and $\mathbf{S}_i$ determine the ellipsoid's rotation and scale. SH represents view-dependent color using 48 coefficients per Gaussian.

\subsection{Differentiable Gaussian Ray Tracing}

In this paper, we use the term \emph{Gaussian Ray Tracing (GRT)} to broadly refer to approaches that optimize a Gaussian scene representation through a differentiable ray-traced renderer~\cite{condor2025dontsplat,moenneloccoz20243dgrt,blanc2025raygauss,byrski2025raysplatsraytracingbased,poirierginter2026gray,xu2026stochastic,vit20263dgaussianacceleratedray}. During training, each iteration goes through three sequential phases: \emph{forward pass}, \emph{backward pass}, and \emph{BVH update}. In the forward pass, a ray is cast from each pixel into the scene, and the contributions of the Gaussians it intersects are blended to produce the pixel color. Testing every ray against millions of Gaussians would be prohibitively expensive, so ray tracers organize the scene in a \emph{Bounding Volume Hierarchy (BVH)}: a tree of nested bounding boxes, where a ray descends only into the boxes it intersects and thereby skips most of the scene. Because rendering is differentiable, the loss against the ground-truth image is back-propagated in the backward pass to obtain gradients for each Gaussian's position, covariance, opacity, and color, which the optimizer uses to update the Gaussians. These updates move and resize Gaussians, and periodically add or remove them, so the bounding boxes stored in the BVH no longer match the scene. The BVH must therefore be updated before the next iteration can trace rays correctly.

\subsection{3DGRT Pipeline and Ray Tracing Algorithm}
\label{sec:3dgrt}

We build on 3DGRT~\cite{moenneloccoz20243dgrt}, a representative and widely used GRT method with an open-source implementation~\cite{3dgrutrepo}.
It follows a pipeline similar to that of 3DGS where, instead of splatting, it uses ray tracing for rendering the scene. BVH traversal, accelerated by the GPU's ray-tracing hardware, passes each hit (aka intersecting Gaussian) to the any-hit shader in no guaranteed order. The shader therefore maintains the $k=16$ nearest hits in sorted front-to-back order within the ray payload. A hit is \emph{accepted} only if blending actually uses it: its contribution along the ray must exceed a minimum threshold, and the ray must not have terminated yet. After each batch, the accepted hits are blended in front-to-back order to update the pixel color. 
If further contribution is still required (the ray transmittance has not reached the minimum threshold), traversal restarts with the ray’s search interval starting just beyond the last hit. 3DGRT runs this entire loop of traversal, sorting, and blending inside a single ray-generation kernel, which exits only once its ray has gathered and blended all contributions. We refer to such a kernel as a \emph{megakernel}, following the ray-tracing literature~\cite{10.1145/2492045.2492060}. Unlike the operator-fusing megakernels of ML systems, the term here only denotes that the entire per-ray loop runs in one kernel.
The resulting pixel colors form the rendered image, which is then compared with the ground-truth image to compute the loss. 

3DGRT stores the farthest accepted hit distance for each pixel during the forward pass. This hit distance is used to bound the retracing during the backward pass to recover the same contributing Gaussians and compute gradients for their parameters. These gradients are summed together in global memory using atomic operations, and used to update Gaussian parameters. Since these updates can modify Gaussian's location, size and shape, 3DGRT rebuilds the BVH at the end of every training iteration. 
Like 3DGS, 3DGRT also periodically densify, prune and reset the opacity of Gaussians at scheduled intervals during training.

\subsection{Where is the Bottleneck in 3DGRT Training?}
\label{sec:bottleneck}

\begin{figure}[t]
    \centering
    \includegraphics[width=\columnwidth]{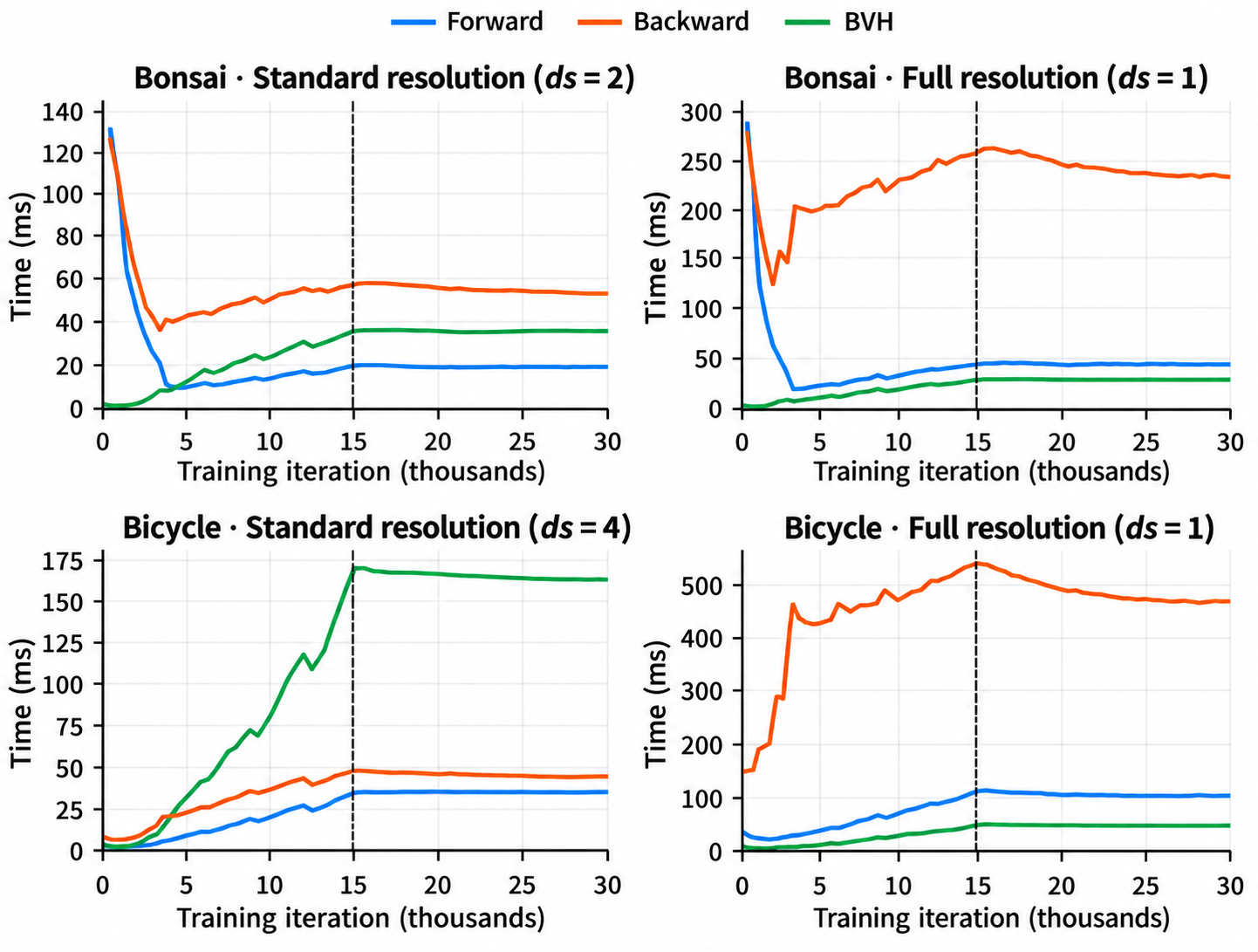}
    \caption{Breakdown of 3DGRT training time for the Bonsai and Bicycle scenes across training stages and image resolutions.}
    \label{fig:breakdown}
\end{figure}

Figure~\ref{fig:breakdown} shows the training-time breakdown for Bonsai and Bicycle from Mip-NeRF 360~\cite{barron2022mipnerf360unboundedantialiased}.  
In particular, Figure~\ref{fig:breakdown} shows that for full resolution, the backward pass dominates the training time across all iterations, making it the biggest target for optimization. Even for lower (e.g., 1/2) resolution, it stays the bottleneck for some indoor scenes like Bonsai; however, for outdoor scenes such as Bicycle, at 1/4 resolution, the BVH update takes over to become the bottleneck after a few iterations. In all the workloads, the backward pass contributes considerable time and thus is a major target for our optimization.

This shift in bottleneck can be explained by an increase in the Gaussian population, which we will discuss further in Opportunity 3 of \S\ref{sec:2.5}. Effectively, with lower resolution, BVH time increases because of an increase in the Gaussian population (refer to Table~\ref{tab:gp}), and at the same time, the backward-pass cost decreases because fewer rays are processed, and the larger spacing between neighboring rays reduces overlap in their Gaussian contributions and therefore atomic operation contention.

\begin{table}[t]

\small
\resizebox{\columnwidth}{!}{
\begin{tabular}{lccccccc}
\toprule
Resolution/Scene & Bonsai & Room & Counter & Kitchen & Bicycle & Garden & Stump\\
\midrule
Standard & 1.5 & 1.2 & 1.2 & 1.4 & 6.8 & 4.5 & 8.5 \\
Full & 1.4 & 1.0 & 1.0 & 1.2 & 2.2 & 3.2 & 1.9 \\
\bottomrule
\end{tabular}
}
\caption{Gaussian population (in millions) at the final trained checkpoint for scenes in the Mip-NeRF 360 dataset~\cite{barron2022mipnerf360unboundedantialiased}. \emph{Standard} denotes the commonly used downsampled resolution (1/2 for indoor and 1/4 for outdoor scenes); \emph{Full} denotes the original resolution.}
\label{tab:gp}
\end{table}

\subsection{Opportunities and Challenges}
\label{sec:2.5}

We identify three opportunities to reduce the cost of 3DGRT training. Each raises a challenge, which we address in Sections~\ref{sec:3.2}, \ref{sec:3.3}, and \ref{sec:3.4}, respectively.

\paragraph{Opportunity 1: Reuse forward-pass hits instead of retracing.} 3DGRT retraces rays during backpropagation to recover the Gaussian hit sequence required for gradient computation. The choice to recompute rather than reuse is inherited from Gaussian-splatting-based approaches where recomputing projections is very fast. However, this is not true for ray tracing based approaches because of costlier BVH traversal and intersection processing. Reusing forward-pass intersections would eliminate this redundant traversal altogether. 

\emph{Challenge: enabling reuse efficiently is non-trivial.}
Recording hits decreases work in the backward pass but increases work in the forward pass. It increases memory pressure in the megakernel and can slow the forward pass enough to offset the backward-pass savings. 

Deciding on the amount of memory to allocate for a pixel is also non-trivial. Although the number of accepted Gaussian hits per pixel is theoretically bounded as shown in Appendix~\ref{sec:appendixa}, directly storing only accepted hits leads to irregular memory accesses. Rays within the same subgroup can accept different numbers of hits, so subsequent writes from neighboring threads diverge to distant addresses, reducing memory coalescing and degrading performance. 

Storing every gathered hit produces a more regular memory access because the write location for each traversal step is known in advance. This allows the archive to be arranged so that neighboring rays write to nearby addresses, improving memory coalescing. The challenge, however, is capacity: the number of gathered hits per ray is not known beforehand and varies significantly across pixels and training iterations. For example, on the Bonsai scene, the maximum hit count across all pixels spikes to roughly 2,700 at iteration 3,001, while remaining closer to 1,000 for most other iterations.

Forward-state reuse therefore requires careful design of the tracing pipeline, storage layout, and hit-capacity policy rather than simply adding a hit buffer.

\begin{figure}[t]
    \centering
    \includegraphics[width=0.9\columnwidth]{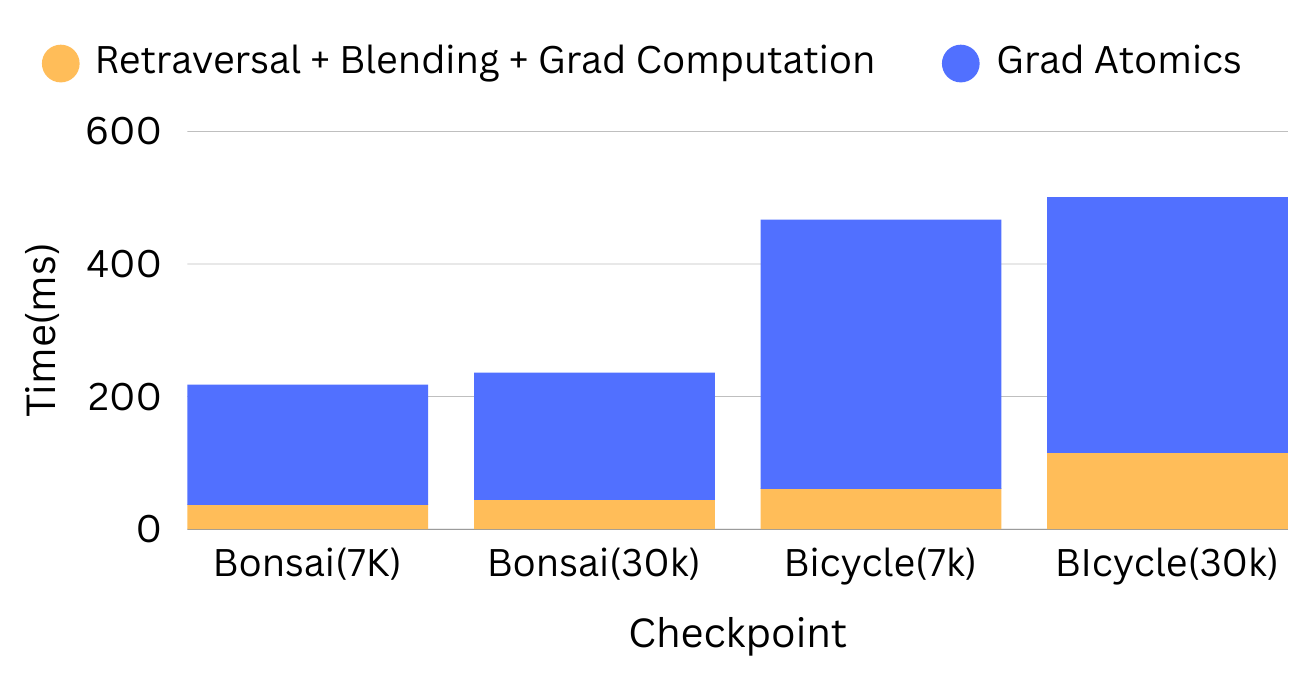}
    \caption{Breakdown of 3DGRT backward-pass time into retraversal, blending, and gradient computation versus gradient atomics, for Bonsai and Bicycle at full resolution.}
    \label{fig:back-split}
\end{figure}

\paragraph{Opportunity 2: Reduce global atomics for gradient accumulation} Using atomics for gradient accumulation results in significant cost for the backward pass. 
To isolate its effect, we compare a backward pass with the global atomics removed while retaining everything else, using the known traversal ranges for each pixel computed during the forward pass against the complete backward pass. The difference in these times is directly attributed to global atomics as shown in Figure~\ref{fig:back-split}, which presents results for the Bonsai and Bicycle scenes at full resolution.

\emph{Challenge: How to reduce gradient atomics?} If rays contributing to the same Gaussian at a given step can be identified, their gradients can be combined in faster on-chip memory before issuing global atomic updates. However, for our Vulkan-based implementation, identifying such rays requires communication across invocations within a warp (a \emph{subgroup} in Vulkan), and Vulkan does not guarantee which pixels share a subgroup.



\paragraph{Opportunity 3: Avoid rebuilding the BVH every iteration.}
\label{par:rebvh}
3DGRT represents each Gaussian using an icosahedral proxy which bounds the Gaussian primitive tightly (hence, reducing false-positive hits) using 20 triangular faces. This allows traversal to exploit hardware-accelerated ray--triangle intersection. This improves traversal efficiency, but each Gaussian adds 20 triangle primitives to the acceleration structure. The issue is more pronounced for outdoor scenes at lower resolutions, where the Gaussian count can increase substantially, unlike indoor scenes where it remains relatively stable (Table~\ref{tab:gp}). For example, Bicycle reaches about $6.8$M Gaussians, corresponding to $\sim136$M triangle primitives, making a full BVH rebuild at every iteration a significant systems cost.

\emph{Challenge: can one refit an existing BVH as Gaussians  change?} Alternative BVH organizations based on instancing can reduce this geometry footprint at the cost of traversal time. For a direct comparison with 3DGRT, however, we retain its original icosahedral representation and instead avoid rebuilding the BVH at every iteration. On the other hand, the cheaper alternative of refitting cannot accommodate changes in the number of Gaussians, and repeated refits gradually degrade traversal efficiency.

\begin{figure*}[t]
    \centering
    \includegraphics[width=0.72\textwidth]{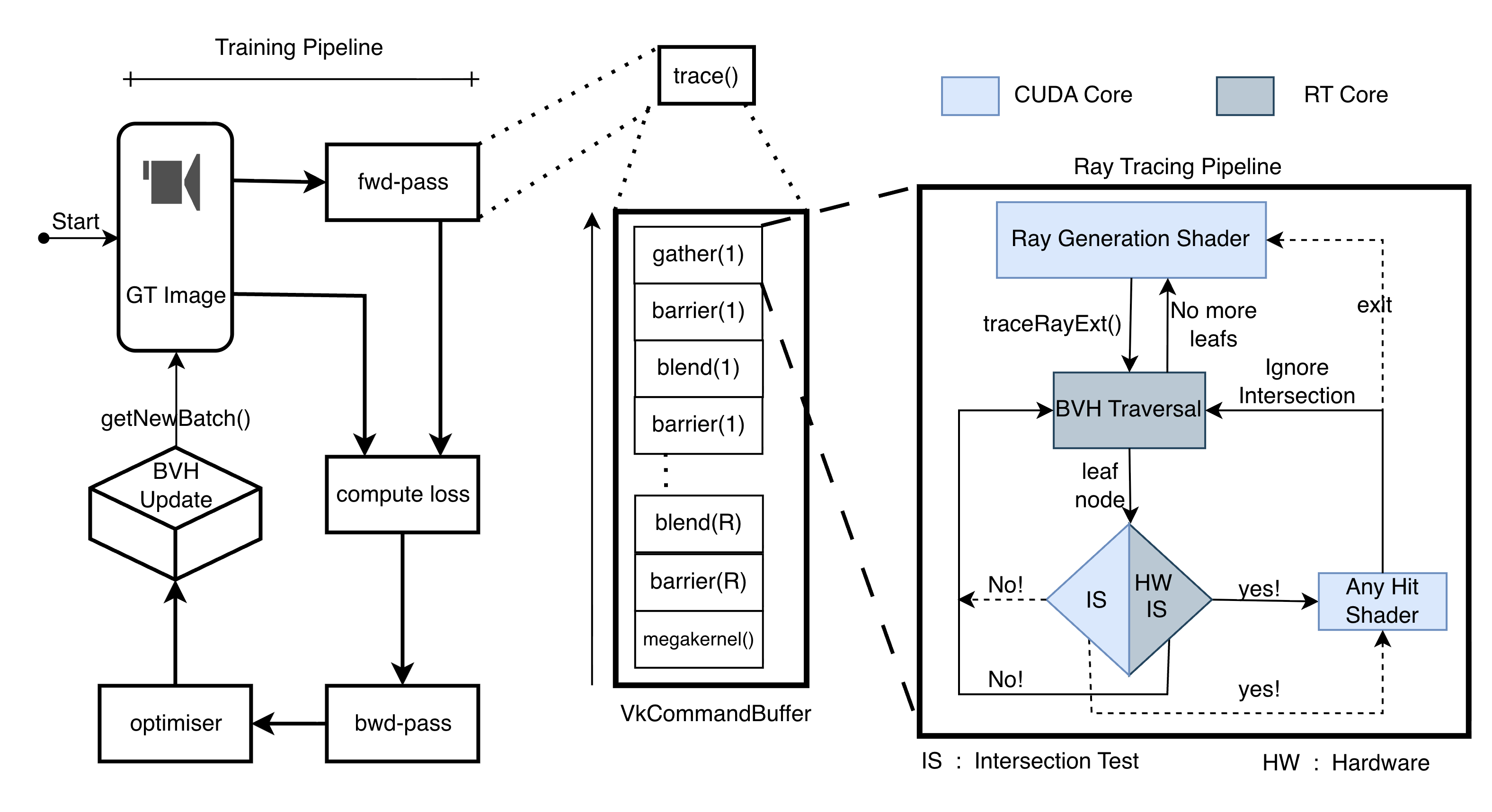}
    \caption{Overview of our Vulkan Training and Ray Tracing pipeline.}
    \label{fig:vulkan_pipeline}
\end{figure*}

\section{System Design}
\label{sec:impl}

\subsection{Training and Ray Tracing Pipeline}
Our system follows the general pipeline of GRT with the rendering formulation of 3DGRT. 
In particular, our optimizations do not alter the mathematical computation performed by 3DGRT. In other words, our system gathers and blends the same Gaussians as 3DGRT for the same scene and camera view for every pixel.

Our training pipeline differs from that of 3DGRT in how information is carried from the forward to the backward pass. Instead of retracing rays during backward computation, as discussed in Opportunity~1 of Section~\ref{sec:2.5}, we preserve forward-pass hit information for reuse. To make this practical, we introduce \textbf{Adaptive Multi-Round Traversal}(\S\ref{sec:3.2}), which combines a round-based traversal pipeline with an adaptive mechanism that adjusts the number of rounds across training iterations.

During the forward pass, each pixel maintains a hit archive storing a fixed number of Gaussians. During the backward pass, each pixel processes its archived hits sequentially and computes the corresponding gradients. To efficiently accumulate these updates without race conditions, we introduce \textbf{ Fast Gradient Accumulator}(\S\ref{sec:3.3}).

Finally, the optimizer updates the Gaussian parameters using the computed gradients. As Gaussian positions and scales change during training, the acceleration structure must also be updated to reflect the new spatial distribution. To reduce this maintenance cost, we introduce \textbf{Efficient BVH Update Strategy} (\S\ref{sec:3.4}).

Figure~\ref{fig:vulkan_pipeline} gives an overview of the resulting pipeline. We discuss each of these three components in the following subsections.

\subsection{Adaptive Multi-Round Traversal}
\label{sec:3.2}
To reuse hits without slowing the forward pass (Opportunity~1), we restructure tracing so that every hit can be recorded cheaply. Motivated by prior work showing the drawbacks of megakernels for ray tracing~\cite{10.1145/2492045.2492060} and by the best practice of keeping ray-generation shaders lightweight, we split the megakernel into two kernels: gather and blend. This splitting reduces memory pressure, but introduces a new execution challenge. Ray traversal
cannot proceed independently of blending: after each batch of hits is gathered, their contributions must be blended to update the ray
transmittance and determine whether further traversal is required. We refer to
one traversal followed by one blending stage as a \emph{round}. The forward
pass therefore becomes a sequence of dependent gather--blend rounds.

As gather and blend are different kernels, the accumulation of a round's blend must be saved in global memory for the next blend to further accumulate. In particular, the following state must be preserved between gather and blend: T (remaining ray transmittance), accumulated radiance (R, G, B), hitDist (where the next gather should start), cursor (represents whether the ray is live or terminated), accumulated depth and hit count.


\paragraph{Hit archive.} Each pixel stores every gathered hit, not only the accepted ones, in a hit archive of 512 slots. This is so that neighboring rays can write to nearby addresses (Section~\ref{sec:implementation}). Because we preallocate 512 slots per ray, a few rays may gather more than 512 Gaussians. For these rays, the backward pass retraverses the BVH to recover the remaining hits and computes their gradients following 3DGRT.

\paragraph{Adaptive round budget.} A fixed round budget $R$ causes two problems. First, for rays that require more than $R$ rounds (which we call \emph{long rays}), a fixed $R$ would terminate the ray before gathering considerable contributions, diverging from the 3DGRT rendering formulation. Second, as training progresses, many rays (which we call \emph{short rays}) terminate within only a few rounds. Even though they no longer traverse, they must still read their state in every remaining round only to determine that they should exit, wasting time until all $R$ rounds complete.
With a fixed value such as $R=16$, this overhead can become substantial once most rays terminate after only a few rounds.

To account for \emph{long rays}, we therefore execute the forward pass in two phases. The first phase performs a bounded number of lightweight traversal--blend rounds. Short rays finish within the first $R$ rounds, after which the remaining long rays are handled by a fallback megakernel, which continues the 3DGRT traversal until their termination criterion is satisfied. The fallback is launched over all pixels, but already terminated rays immediately exit, leaving only the small tail of long-running rays to perform additional work.

As the Gaussian representation becomes denser and better fits the scene, rays generally accumulate opacity more quickly and tend to terminate earlier. To account for \emph{short rays}, we therefore adapt the round budget from the behavior observed in the previous iteration. Let $p_t$ denote the fraction of rays that remain active after reaching the round budget $R_t$. We initialize $R_0=16$ and update

\begin{equation}
R_{t+1} =
\begin{cases}
R_t + 1, & p_t \geq 1\%, \\
\max(1,R_t - 1), & p_t < 0.25\%, \\
R_t, & \text{otherwise}.
\end{cases}
\end{equation}

\begin{figure}[t]
    \centering
    \includegraphics[width=0.8\columnwidth]{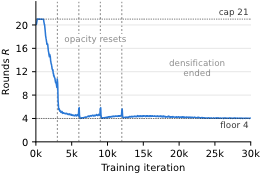}
    \caption{Adaptive round budget over training iterations for the Bonsai scene at full resolution.}
    \label{fig:rounds}
\end{figure}

Changing the budget by only one round at a time prevents transient behavior
from causing large scheduling changes between consecutive iterations. If too
many rays exhaust the current budget, we increase $R$; if almost all rays
terminate earlier, we decrease it. In our experiments, the initial budget of
$R=16$ gradually decreases to approximately $R=4$ (Figure~\ref{fig:rounds}), while automatically increasing again when training events temporarily require
longer traversal. 

\begin{figure}[t]
    \centering
    \includegraphics[width=\columnwidth]{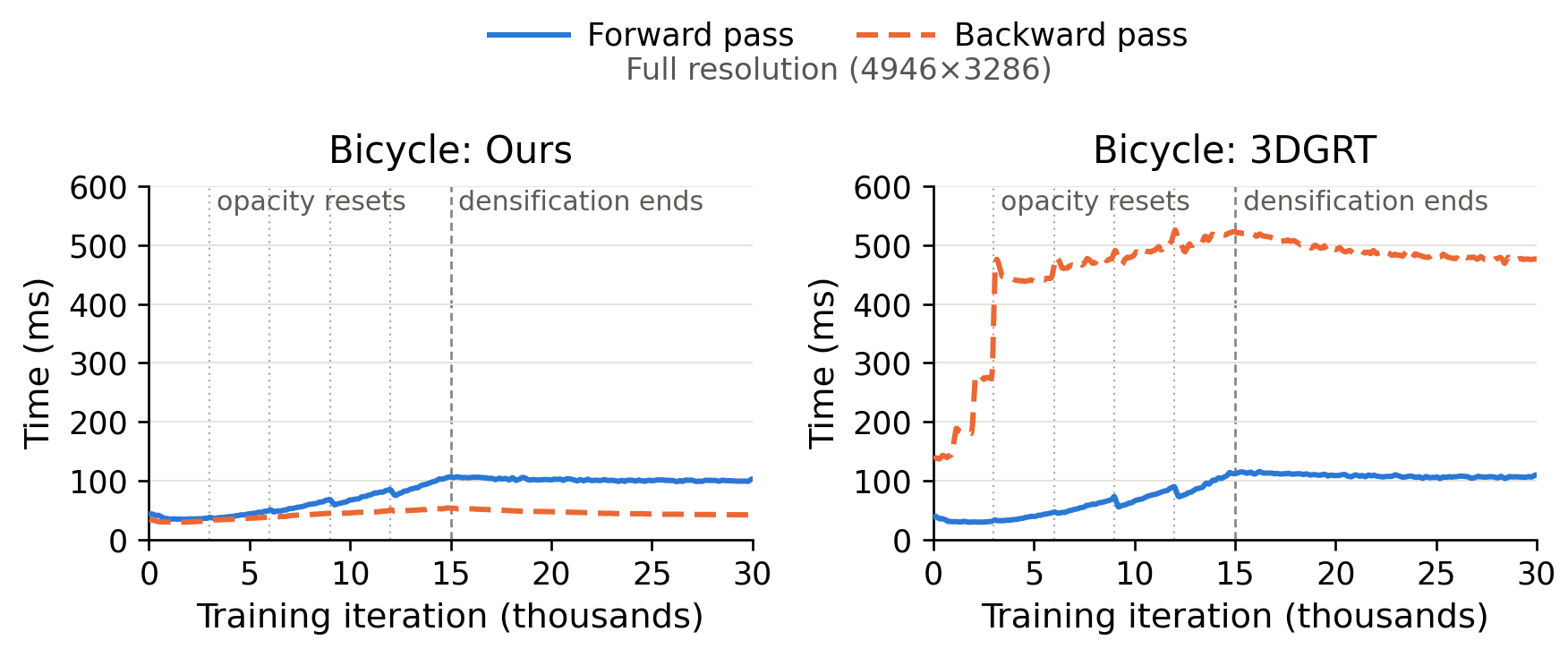}
    \caption{Forward-pass and Backward-pass time of \method and 3DGRT over training iterations.}
    \label{fig:fwdpass}
\end{figure}

This design makes it possible to record hits during the forward pass without additional overhead and without capping the number of rounds per ray, eliminating retraversal in the backward pass. Figure~\ref{fig:fwdpass} shows speedup in backpass with little overhead in forward pass.

\subsection{Fast Gradient Accumulator}
\label{sec:3.3}
As discussed in Opportunity 2 (Section~\ref{sec:2.5}), gradient accumulation is one of the major bottlenecks. We reduce these atomics by aggregating contributions within each subgroup before writing to global memory.

Our backward compute kernel processes a workgroup corresponding to an 8 × 8 tile of adjacent pixels, comprising two 32-lane subgroups on our target NVIDIA GPUs. Neighboring pixels often contribute to the same Gaussians, creating opportunities for aggregation. Although Vulkan does not generally guarantee the mapping of local invocation indices to subgroup lanes, all invocations remain within the same pixel tile. Our method therefore exploits tile-level spatial locality without requiring a particular subgroup layout.

\begin{figure}[t]
    \centering
    \includegraphics[width=\columnwidth]{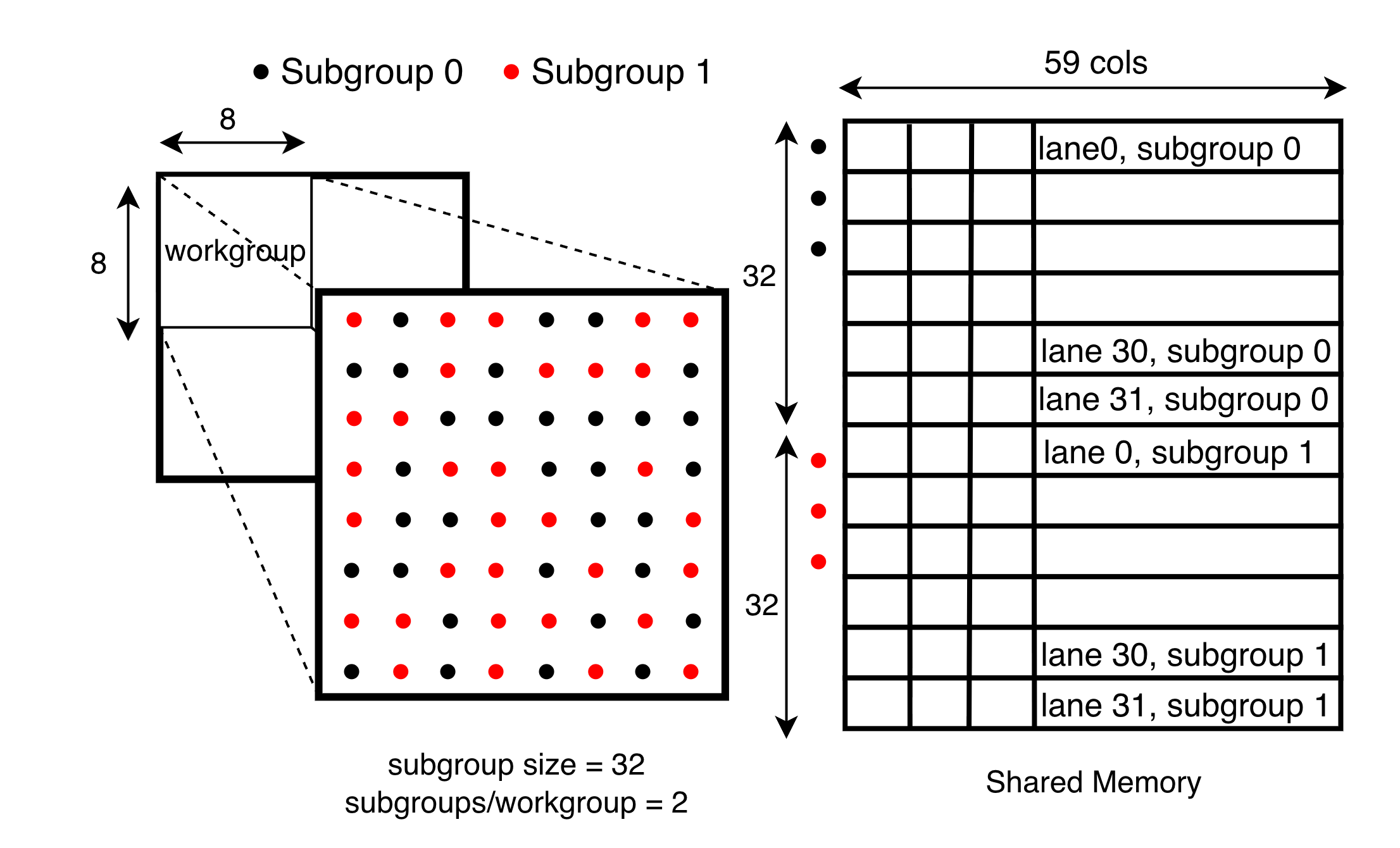}
    \caption{Per-lane Gaussian gradients are staged in shared memory, and the accumulation work is divided among lanes.}
    \label{fig:sharedmem}
\end{figure}

The kernel processes the cached Gaussian lists one entry at a time. At each list position, lanes with valid entries compute their gradient contributions and stage them in separate shared-memory rows (Figure~\ref{fig:sharedmem}) indexed by \textit{(subgroupID, laneID)}. Each row contains 59 gradient components, requiring approximately 14.75\,KiB of shared memory per 64-invocation workgroup.

\begin{algorithm}[t]
\caption{Subgroup-Aggregated Gradient Accumulation}
\label{alg:grad_accum}
\begin{algorithmic}[1]

\REQUIRE Gaussian ID $g_i$ and gradient $\mathbf{d}_i$
\STATE $A \gets$ active lanes \COMMENT{Find lanes having accepted hit}

\WHILE{$A \neq 0$}
    \STATE $l \gets \textsc{FirstActiveLane}(A)$

    \COMMENT Get gId of lane l and mask lanes with same gId
    \STATE $g \gets \textsc{Broadcast}(g_l, l)$ 
    \STATE $M \gets \textsc{Ballot}(A \land (g_i = g))$ 

    \COMMENT Distribute 59 gradient components to accumulate among all lanes
    \FOR{$d \gets \textsc{LaneID}()$ \textbf{to} $D-1$ \textbf{step} 32}
        \STATE $s \gets 0$
        \FOR{all $j \in M$}
            \STATE $s \gets s + \mathbf{d}_j[d]$
        \ENDFOR
        \COMMENT One atomic
        \STATE $\textsc{AtomicAdd}(\nabla\theta_g[d], s)$
    \ENDFOR

    \STATE $A \gets A \setminus M$
\ENDWHILE

\end{algorithmic}
\end{algorithm}

Within each subgroup, we identify valid lanes (those with accepted Gaussian hits) using a ballot and partition their contributions by Gaussian ID. For each distinct Gaussian, lanes cooperatively reduce the corresponding rows, distributing gradient components across lanes: lane $\ell$ accumulates component $\ell$ and, when present (i.e., if $\ell + 32 < 59$), component $\ell+32$. Each accumulated component is then written to global memory with a single atomic addition. This distributes the reduction across the subgroup rather than assigning the entire gradient vector to one lane. Algorithm~\ref{alg:grad_accum} summarizes the procedure, and Figure~\ref{fig:fwdpass} compares the resulting backward pass with that of 3DGRT.

\subsection{Efficient BVH Update Strategy}
\label{sec:3.4}
BVH refitting offers a cheaper alternative to rebuilding: it updates only the node bounds and preserves the existing hierarchy, without reconstructing any nodes. However, repeated refitting can degrade traversal efficiency as the geometry moves away from the spatial partitioning established during construction. Refitting also cannot accommodate changes in primitive count, requiring a rebuild when densification or pruning modifies the Gaussian population. 

\begin{figure}[t]
    \centering
    \includegraphics[width=\columnwidth]{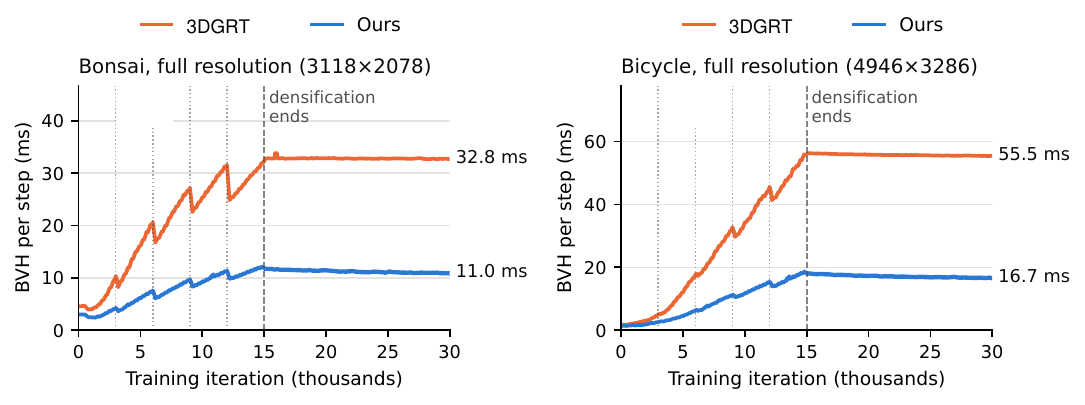}
    \caption{BVH update time per step of \method and 3DGRT over training iterations.}
    \label{fig:bvh}
\end{figure}

Hybrid rebuild/refit strategies are common for dynamic scenes~\cite{lauterbach2006rtdeform,kopta2012bvhupdates}, but to the best of our knowledge they have not been explored for 3DGRT training. We therefore adopt a hybrid strategy that combines refitting with selective rebuilding. During the first 15,000 iterations, when densification is active, we rebuild at densification and pruning events and refit at all other iterations. After densification ends, the Gaussian population remains fixed, but continued parameter updates can still degrade BVH quality. We therefore rebuild every 16 iterations and refit in between, balancing maintenance cost against traversal efficiency (Figure~\ref{fig:bvh}).

\section{Implementation}
\label{sec:implementation}

\paragraph{Command Buffers.} Launching kernels separately for each of the $R$ traversal rounds introduces repeated host-side submission overhead. Vulkan allows multiple GPU commands to be recorded into a command buffer and submitted together to a \texttt{VkQueue}, reducing this overhead. However, dependencies between these commands are not inserted automatically; required execution and memory ordering must be explicitly specified using synchronization primitives such as pipeline barriers.

\paragraph{Shader Launch Parameters.}
For the blending and backward compute shaders, we use an $8\times8$ workgroup layout. The ray-tracing gather stage differs because Vulkan ray-tracing shaders do not use workgroups, instead, shader invocations are organized into subgroups by the implementation.











\begin{figure*}[t]
    \centering
    \includegraphics[width=\textwidth]{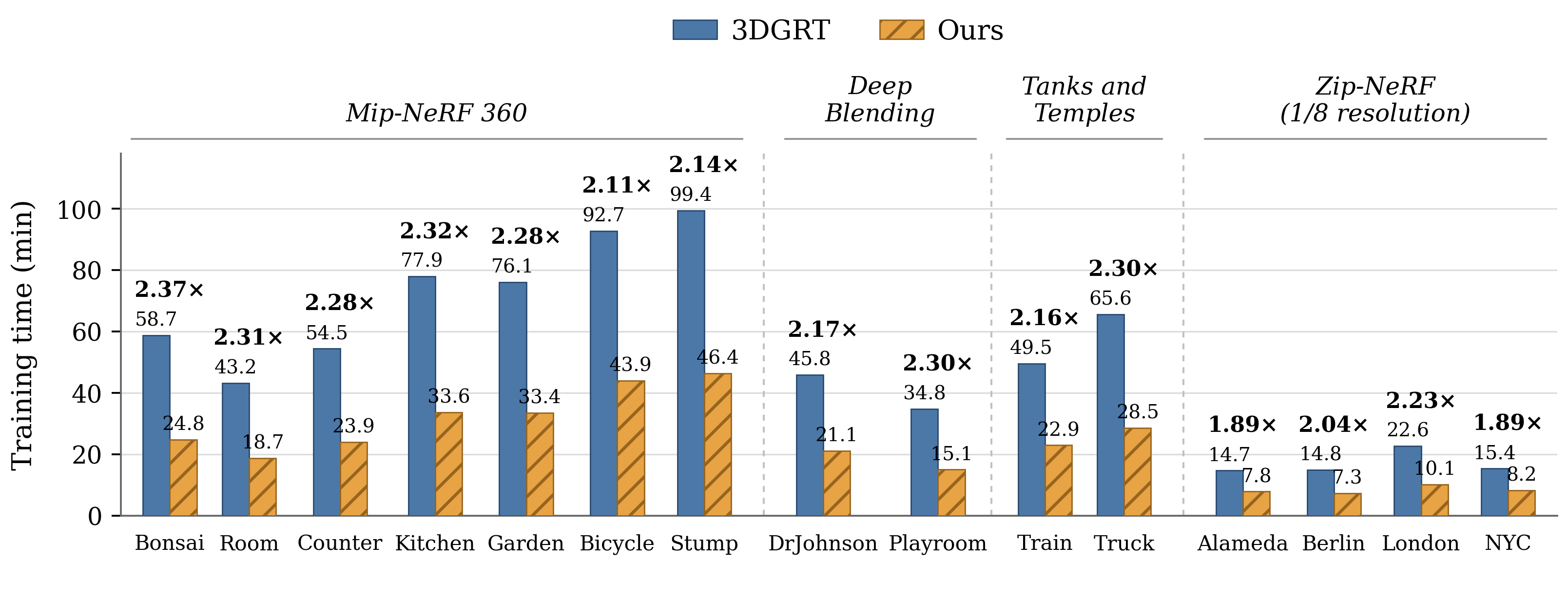}
    \caption{End-to-end training performance at Standard resolution.}
    \label{fig:e2et}
\end{figure*}

\paragraph{Any-hit and $K$ value.}
In 3DGRT, each retained hit stores both its distance and Gaussian ID in the ray payload, although only the distance is needed during hit selection. We move Gaussian IDs to global memory and maintain an unsorted distance buffer, tracking the farthest hit for replacement and sorting only before blending. These changes reduce payload pressure and increase the optimal $K$ from 11 to 24 in our Vulkan implementation. A larger $K$ reduces traversal rounds but increases per-round traversal and sorting costs. Refer Appendix~\ref{app:anyhit} for further details.

\paragraph{Memory Layout.}
Throughout our design, we organize stored state to favor coalesced memory accesses. For hit replay, nearby pixels should access Gaussian hits from nearby memory locations. We therefore use a slot-major layout, where each column corresponds to a pixel/ray and each row corresponds to one of the 512 hit slots. As a result, the same hit slot for neighboring pixels is stored contiguously, enabling more coalesced accesses during both recording and replay.

\section{Experiments and Evaluation}

We evaluate the following implementations:

\begin{itemize}[leftmargin=*]
    \item \textbf{3DGRT Baseline}: OptiX implementation\footnote{For reproducibility, all 3DGRT experiments use the official implementation from \url{https://github.com/nv-tlabs/3dgrut/tree/e6f4552}.} of the original 3DGRT~\cite{moenneloccoz20243dgrt}, which follows the same densification strategy as 3DGS and uses a density learning rate of 0.09. During inference, it uses density clamping and a transmittance threshold of 0.03.
    
    \item \textbf{\method (Ours)}: our Vulkan implementation with all the optimizations described above, using exactly the same rendering and training configuration as 3DGRT. 
\end{itemize}

We do not compare against GRTX~\cite{lee2026grtxefficientraytracing}, the only other Vulkan-based Gaussian ray tracer we are aware of, because its source code is not publicly available. Its proposed hardware extensions would also require architectural simulation rather than execution on existing GPUs.

\paragraph{Experimental Setup.}
Experiments were conducted on an NVIDIA RTX PRO 6000 Blackwell GPU with driver 610.43.02, Vulkan 1.4, and CUDA 12.8.1. We evaluate on Mip-NeRF 360~\cite{barron2022mipnerf360unboundedantialiased}, Tanks and Temples~\cite{Knapitsch2017}, Deep Blending~\cite{DeepBlending2018}, and Zip-NeRF~\cite{barron2023zipnerf}. For Mip-NeRF 360, we use the standard $1/2$ indoor and $1/4$ outdoor resolutions. Following 3DGRT, we report mean PSNR, LPIPS, and SSIM over test views for reconstruction quality and FPS for rendering performance.






\subsection{Performance}

\begin{figure}[t]
    \centering
    \includegraphics[width=\columnwidth]{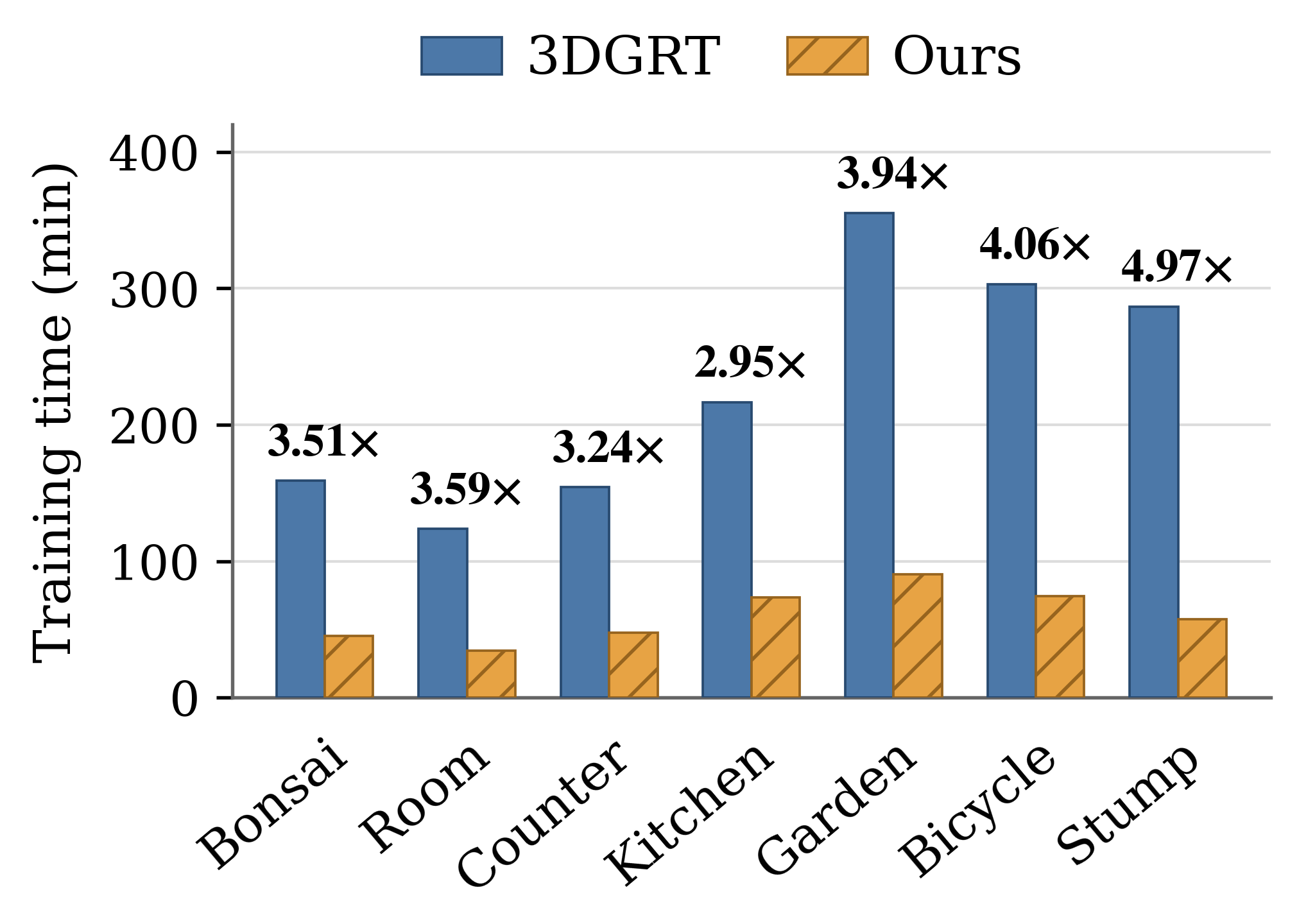}
    \caption{End-to-end training time for MipNeRF360 scenes at full resolution}
    \label{fig:e2e-full}
\end{figure}

\begin{figure}[t]
    \centering
    \includegraphics[width=\columnwidth]{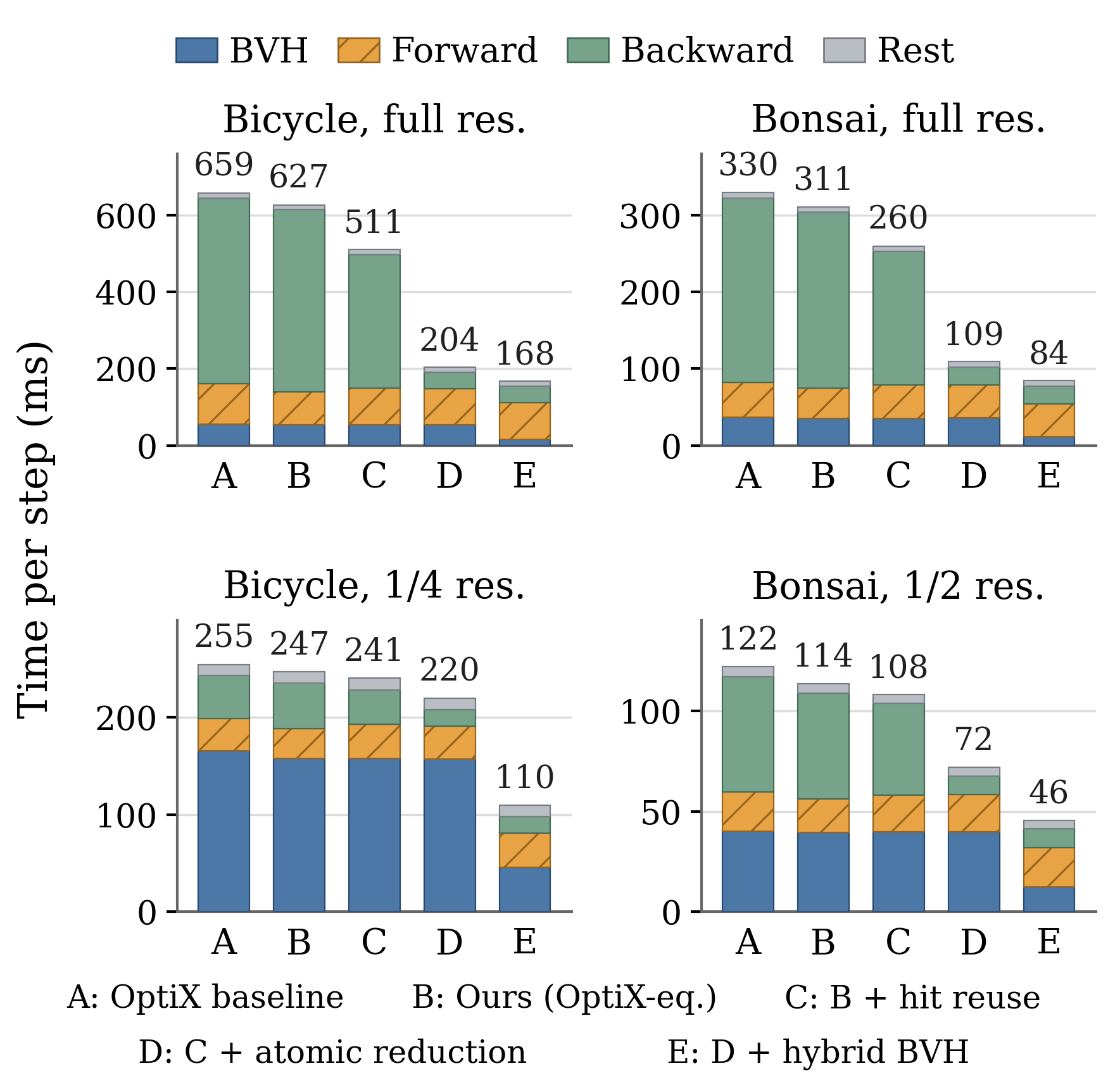}
    \caption{Performance benefit of each of the three proposed optimizations at selected training checkpoints.}
    \label{fig:isolate}
\end{figure}

\paragraph{End-to-end training performance.} Figure~\ref{fig:e2et} reports end-to-end training time for 3DGS, 3DGRT and \method (Ours), along with our speedup over 3DGRT. The speedup is larger at full resolution (Figure~\ref{fig:e2e-full}), mainly because the backward pass becomes much more expensive at high resolution.
Our optimizations provide the largest benefit in the backward pass Figure~\ref{fig:isolate}, translating into greater end-to-end speedups at full resolution. 3DGS does not run on distorted cameras and thus its results are produced.

\begin{figure}[h]
    \centering
    \includegraphics[width=0.8\columnwidth]{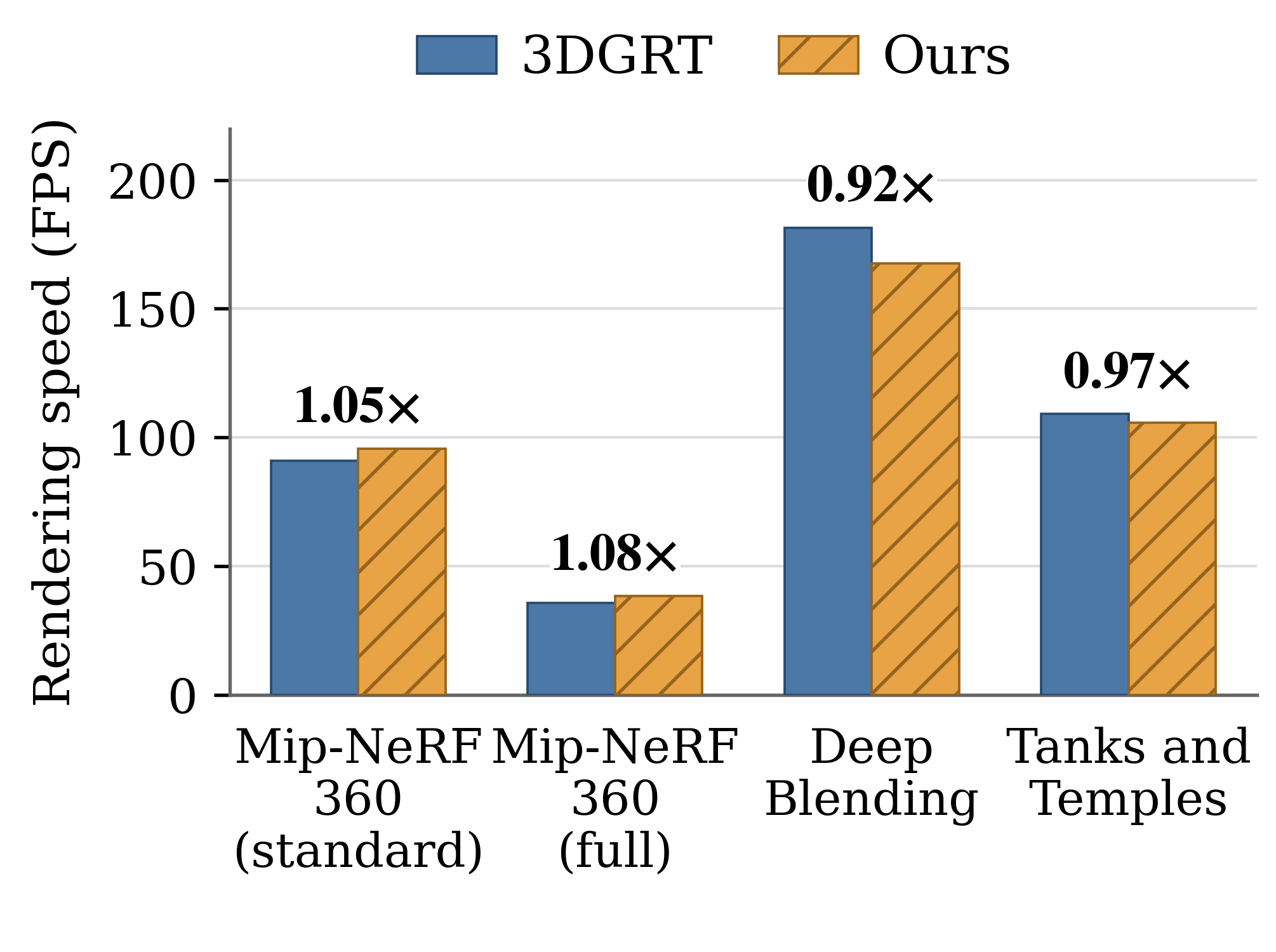}
    \caption{Comparable Inference Time Rendering Speed}
    \label{fig:infer}
\end{figure}

\paragraph{Rendering performance.} At inference, we run the megakernel without hit storage for comparison with 3DGRT (Figure~\ref{fig:infer}). Since inference has no backward pass and BVH update, the megakernel is the better choice there: when memory pressure is low and the computation around traversal does not limit occupancy.












\paragraph{Complex Camera Models}
We evaluate our approach on the Zip-NeRF dataset~\cite{barron2023zipnerf}, which consists of four scenes (Alameda, Berlin, London, and NYC) captured with fisheye cameras and  distorted inputs. Following 3DGRT, we use 1/8 resolution for a fair comparison. 

Because 3DGS relies on projection, it handles such camera models poorly, so we compare only against 3DGRT. Figure~\ref{fig:e2et} shows our training speedups, and Table~\ref{tab:perf} shows that \method achieves comparable rendering quality.








\subsection{Which optimization contributes most?}
Figure~\ref{fig:isolate} shows the effect of each optimization on the individual phases. At full resolution, most of the benefit comes from the atomic reduction (from C to D), which reduces iteration time by 300\,ms. At standard resolution, the gains come from both the atomic reduction (e.g., Bonsai at 1/2 resolution) and the BVH update (e.g., Bicycle at 1/4 resolution). 

\begin{table*}[t]
\centering

\scriptsize

\resizebox{\textwidth}{!}{
\begin{tabular}{lcccccccccccc}
\toprule

&
\multicolumn{3}{c}{\textit{Mip-NeRF 360}} &
\multicolumn{3}{c}{\textit{Deep Blending}} &
\multicolumn{3}{c}{\textit{Tanks and Temples}} &
\multicolumn{3}{c}{\textit{Zip-NeRF}} \\

\cmidrule(lr){2-4}
\cmidrule(lr){5-7}
\cmidrule(lr){8-10}
\cmidrule(lr){11-13}

Method/Metric &
PSNR$\uparrow$ &
LPIPS$\downarrow$ &
SSIM$\uparrow$ &
PSNR$\uparrow$ &
LPIPS$\downarrow$ &
SSIM$\uparrow$ &
PSNR$\uparrow$ &
LPIPS$\downarrow$ &
SSIM$\uparrow$ &
PSNR$\uparrow$ &
LPIPS$\downarrow$ &
SSIM$\uparrow$ \\

\midrule

3DGRT
& 27.88
& 0.241
& 0.851
& 29.45        
& 0.332
& 0.896
& 22.54
& 0.216
& 0.829 
& 24.33
& \textbf{0.257}
& \textbf{0.848} \\

\method (Ours)
& \textbf{28.08}
& 0.241
& \textbf{0.852}
& 29.45
& \textbf{0.330}
& 0.896
& 22.54
& \textbf{0.214}
& \textbf{0.831}
& \textbf{24.34}
& 0.261
& 0.847 \\

\bottomrule
\end{tabular}
}

\caption{Rendering quality of the final converged checkpoints at standard resolution (mean over all scenes in each dataset).}
\label{tab:perf}
\end{table*}



\subsection{Peak GPU Memory Usage}
Table~\ref{tab:peak-mem-std} reports peak GPU memory usage over the entire training run. At standard resolution, the number of Gaussians grows substantially for outdoor scenes, which is reflected in the table: 3DGRT uses around 30\,GiB for these workloads. \method consistently uses more memory than 3DGRT because it stores forward-pass hits, whose footprint scales with image resolution.

\subsection{Reconstruction quality at Inference}

Table~\ref{tab:perf} shows the reconstruction quality by rendering the final model checkpoint on test views. Each entry is the mean over all scenes in the dataset, evaluated at standard resolution following previous work on GRT.

We observe a small difference in mean PSNR on Mip-NeRF 360 despite using the same rendering formulation as 3DGRT. In earlier experiments, we identified an upstream 3DGRT issue where floating-point precision cause the final forward-pass hit to be skipped during backward traversal, because of which we see this difference.

\begin{table}[h]
    \centering
    \scriptsize
    \setlength{\tabcolsep}{4pt}

    \begin{tabular}{lrr}
      \toprule
      Scene/Method & \method (GiB) & 3DGRT (GiB) \\
      \midrule
      Bonsai  & 13.1 & 7.5 \\
      Counter & 10.9 & 5.3 \\
      Kitchen & 13.0 & 5.8 \\
      Room    & 10.0 & 5.4 \\
      Bicycle & 49.0 & 32.6 \\
      Garden  & 30.7 & 19.5 \\
      Stump   & 40.4 & 33.0 \\
      \bottomrule
    \end{tabular}

    \caption{Peak GPU memory during training at standard resolution}
    \label{tab:peak-mem-std}
\end{table}

\section{Related Work}
\label{sec:related}


\paragraph{Gaussian Splatting Systems.}
3DGS~\cite{kerbl20233dgs} established 3D Gaussians as an effective differentiable representation for novel-view synthesis. Subsequent work improves different parts of this pipeline: FastGS~\cite{ren2025fastgstraining3dgaussian} reduces training time, CLM-GS~\cite{zhao2025clm} improves scalability under limited GPU memory, 3DGUT~\cite{wu20253dgut} improves Gaussian projection under non-linear camera models using the unscented transform, and EDGS~\cite{Kotovenko_2026_CVPR} revisits densification and pruning for faster convergence.

\paragraph{Ray-Traced Gaussian Representations.}
Several works render Gaussian or ellipsoidal primitives by ray tracing rather than splatting. RayGauss~\cite{blanc2025raygauss} performs volumetric ray casting over irregularly distributed Gaussian basis functions using a BVH. EVER~\cite{mai2025ever} replaces Gaussians with constant-density ellipsoids whose volume rendering integral can be evaluated exactly, eliminating popping artifacts. Condor et al.~\cite{condor2025dontsplat} use volumetric ray-traced Gaussian primitives to model scattering and emissive media. These works focus on the rendering formulation and its quality, whereas we keep the formulation of 3DGRT fixed and optimize how its training executes on the GPU.

\paragraph{Traversal and BVH Optimizations.}
GRTX~\cite{lee2026grtxefficientraytracing} targets the traversal and intersection cost of Gaussian ray tracing. It transforms anisotropic Gaussians into a normalized unit-sphere representation to simplify intersection tests and reduce acceleration-structure complexity, and additionally proposes hardware support to improve BVH traversal. Its focus is therefore complementary to ours: GRTX primarily optimizes the traversal mechanism, while we study the systems organization of the complete training pipeline.

\paragraph{Optimizing 3DGRT Training.}
GRay~\cite{poirierginter2026gray} explores algorithmic and representation-level changes for faster Gaussian ray-tracing training, including transferring dense initialization strategies from splatting. Stochastic ray tracing~\cite{xu2026stochastic} replaces sorted front-to-back blending with an unbiased Monte Carlo estimator that evaluates only a sampled subset of Gaussians per ray, avoiding sorting altogether. Both approaches gain speed by changing the rendering formulation, while, we preserve the exact formulation of 3DGRT and instead reorganize how its training executes. 3DGART~\cite{vit20263dgaussianacceleratedray} identifies gradient accumulation in the backward pass as a major bottleneck and proposes a primitive-major organization using tile--primitive mappings. Its formulation relies on perspective projection, however, limiting its applicability to more general camera models. Our work also targets backward-pass overhead, but preserves the ray-based formulation and does not assume perspective projection. Outside ray tracing, DISTWAR~\cite{durvasula2023distwar} reduces atomic contention in differentiable rasterization pipelines, including 3DGS, by performing warp-level reduction of gradient updates before issuing global atomics. Our fast gradient accumulator follows a similar principle, but must operate without Vulkan guarantees on how invocations map to subgroups, and aggregates over per-ray replayed hit lists rather than rasterized fragments.

\paragraph{Forward-State Reuse.}
RaySplats~\cite{byrski2025raysplatsraytracingbased} also stores forward-pass Gaussian intersections for reuse during backpropagation, avoiding repeated traversal. However, it assumes a fixed cap on the number of hits per ray and changes the ray termination condition to account for this approximation. Our work makes no such approximation. RaySplats also does not report the performance benefit of this reuse. 

\section{Discussion and Future Work}

  \paragraph{The forward pass is the new bottleneck.}
  In the OptiX baseline the backward pass dominates a training step: on
  \textit{bicycle} at full resolution it takes 484\,ms of a 659\,ms step, about
  $4.6\times$ the forward pass. Hit replay, atomic reduction and hybrid BVH updates
  remove most of that cost. With all three, the backward pass falls to 43\,ms and the
  BVH update to 16\,ms. The forward pass, at 96\,ms, is now more than half of the step
  and over twice the backward pass. Our forward pass is only on par with OptiX's
  (1.02--1.06$\times$ at convergence), so it is where further gains in training time
  must come from. Possible directions include adapting the number of hits gathered
  per round to how deep each ray's march goes, reordering rays~\cite{nvidia2022ser} to improve traversal
  coherence, and reduced-precision shading.

\paragraph{Atomic Reduction} Our optimisation benefits from nearby rays hitting same gaussians. Because Vulkan does not guarantee invocation to pixel mapping, it is possible that nearby rays that hit same gaussians are present in different subgroups, thus giving no benefit of accumulating gradients within a subgroup. Additonally, aggregation does not benefits reflective or refractive secondary rays where divergence increases and shared-Gaussian contention decreases, reducing the benefit of atomic aggregation while increasing traversal and memory-access irregularity. 




\section{Conclusion}

\method reduces the training time of 3DGRT without compromising rendering quality. Our key observation is that a considerable fraction of the training cost lies not in ray traversal itself, but in the computation organized around it. By reusing forward-pass hits through adaptive multi-round traversal, aggregating gradients within subgroups before issuing global atomics, and combining BVH refitting with selective rebuilding, \method substantially accelerates end-to-end training while gathering and blending the same Gaussians as 3DGRT. These techniques rely on general properties of Gaussian ray-tracing training rather than on a particular workload, and our Vulkan implementation is not tied to a single vendor's ray-tracing API.





\bibliography{references}
\bibliographystyle{mlsys2025}

\section*{AI Usage Disclosure}

We used OpenAI's ChatGPT and Anthropic's Claude as assisting tools during this work. They were used for language editing, rephrasing, manuscript organization, LaTeX formatting, technical discussion, literature and documentation lookup, and code generation and debugging assistance. Generative AI tools were also used to help create or edit the appearance a small number of figures, while all underlying experimental data were produced by the authors. The technical contributions, system design, implementation, experiments, measurements, analysis, and conclusions were developed and validated by the authors.

\clearpage
\appendix

\section{Theoretical case bound on number of accepted hits per ray in 3DGRT}
\label{sec:appendixa}
The 3DGRT setup provides a theoretical upper bound through its transmittance termination
criterion. Training terminates a ray once its transmittance falls below
$\tau = 0.001$. 

The longest traversal occurs when each accepted hit lets as much light through as possible.

An accepted Gaussian must have minimum opacity of $\alpha_{\min}=1/255$. In the worst case, where every accepted Gaussian
blocks exactly this minimum amount, the transmittance after $n$ hits is $(1-\alpha_{\min})^n$, and the ray remains active only while

\begin{equation}
    (1-\alpha_{\min})^n \geq \tau .
\end{equation}

Thus, the number of accepted Gaussians before termination
satisfies

\begin{equation}
    n \leq
    \frac{\log(0.001)}{\log(\frac{254}{255})}
    \approx 1758.02 .
\end{equation}

Therefore, approximately 1759 accepted Gaussian contributions are sufficient to terminate a ray in the worst case.

\section{Any-Hit Implementation and $K$ Value}
\label{app:anyhit}

\paragraph{Hit storage and payload usage.}
In 3DGRT, each retained hit is represented by two values in the ray payload: its distance and its Gaussian ID. For a buffer of $K=16$ hits, these occupy 32 payload entries. During traversal, hit distances are repeatedly read to compare incoming hits with those already retained. Gaussian IDs, however, are only needed after traversal to identify the Gaussians to be processed. Consequently, half of the payload entries hold values that must be preserved but are not needed for hit selection.

Our implementation separates the storage of these two values. We retain hit distances in the payload and store the corresponding Gaussian IDs in global memory. Each retained hit has a slot index that associates its distance with its ID. When a hit is inserted or replaced, we update the distance and the corresponding ID at that slot. Moving IDs out of the payload frees space for additional hit distances, allowing more hits to be gathered in a single traversal. This trades additional global-memory accesses for reduced payload storage requirements.

\paragraph{Maintaining an unsorted hit buffer.}
We also avoid maintaining a fully sorted hit buffer during traversal. While free slots remain, incoming hits are appended to the next available slot. Once the buffer contains $K$ hits, we only need to determine whether a new hit belongs among the $K$ closest hits encountered so far. This requires comparing its distance with the largest distance currently stored, rather than with every entry in sorted order.

We therefore maintain two additional values: the maximum stored distance, \texttt{maxDistance}, and the index of the slot containing it. If an incoming hit is farther away than the current farthest hit, it is discarded. If it is closer, it replaces the farthest hit in that slot, and the corresponding Gaussian ID is updated in global memory.

After a replacement, the previous maximum is no longer valid. Because the buffer is unsorted, we scan the retained distances to find the new maximum and its index. Hits that do not replace an entry require no such scan. This organization avoids shifting entries to maintain sorted order during traversal, at the cost of recomputing the maximum whenever the farthest hit is replaced. Once traversal finishes, the retained hits are sorted by distance before front-to-back blending.

\paragraph{Effect on the optimal buffer size.}
These changes allow us to retain more hits within the available payload space. In our initial Vulkan implementation, which followed the original 3DGRT buffer organization, the best-performing value was $K=11$. After changing the storage and buffer organization, the best-performing value increased to $K=24$. Although moving Gaussian IDs out of the payload frees approximately half of the original per-hit payload storage, the best-performing $K$ is determined by execution costs as well as storage capacity.

\paragraph{Tradeoff between traversal rounds and per-round cost.}
Gathering more hits per traversal can reduce the number of gather--blend rounds needed by a ray. For a ray that processes the same sequence of contributions, increasing $K$ allows each round to process a larger batch. However, fewer rounds do not necessarily imply lower execution time, because the cost of each round also changes.

First, $K$ affects traversal pruning. Once the buffer is full, the distance of the farthest retained hit can be used to tighten the traversal range. A smaller buffer fills sooner and can establish a tighter distance bound earlier, allowing more distant hits to be excluded. A larger buffer generally retains a farther-reaching set of hits, so traversal may explore more hits before it finishes.

Second, blending determines whether the ray has accumulated enough opacity to terminate. Larger batches may gather hits that are ultimately unnecessary because the ray reaches its termination threshold partway through the batch. Smaller batches allow this termination condition to be checked more frequently, but require more traversal rounds.






\begin{figure*}[p]
    \centering
    \includegraphics[width=\textwidth]{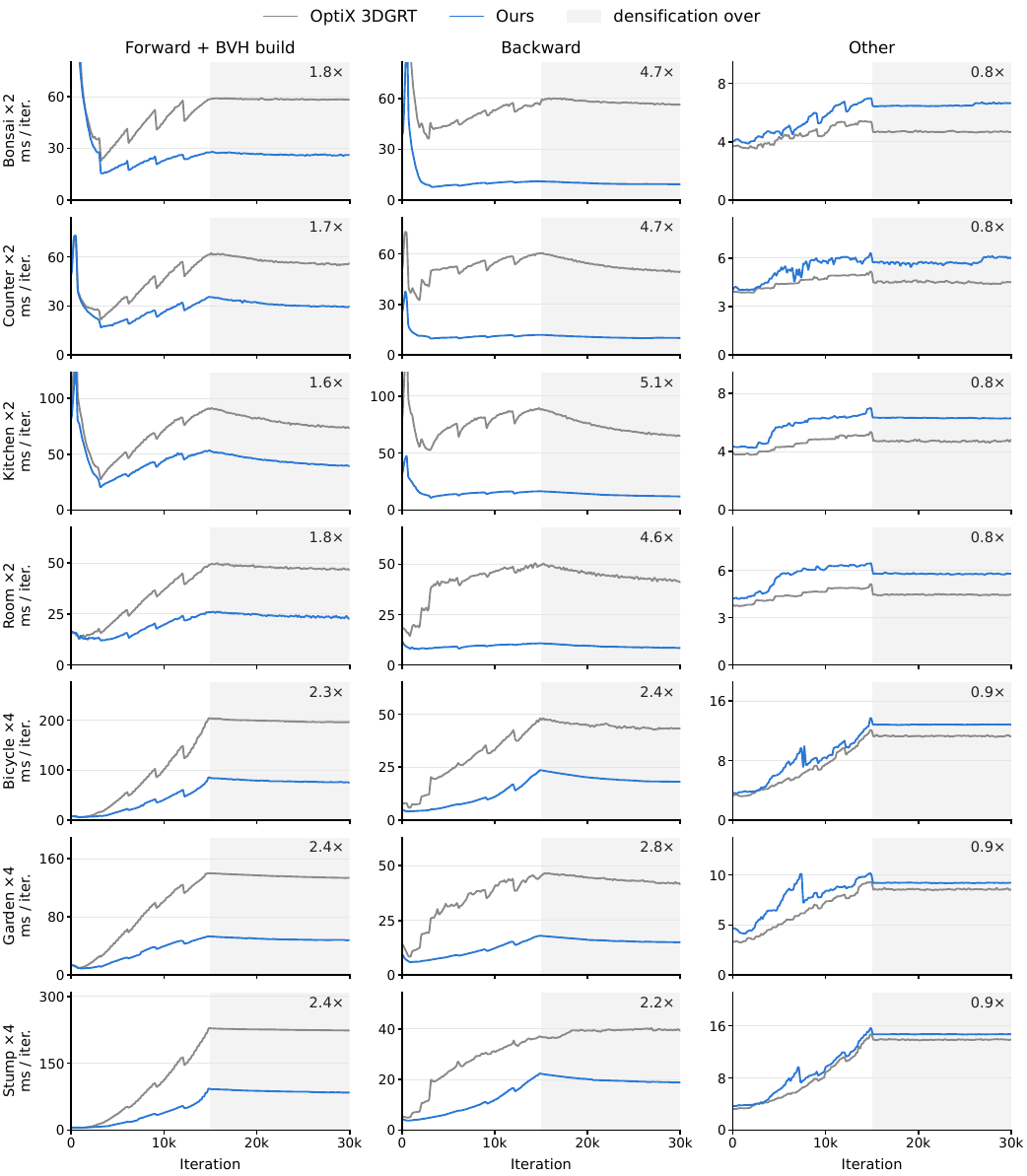}
    \caption{Per-phase time per training iteration, Mip-NeRF~360 at standard resolution.}
    \label{fig:phases-std}
  \end{figure*}
  
  \begin{figure*}[p]
    \centering
    \includegraphics[width=\textwidth]{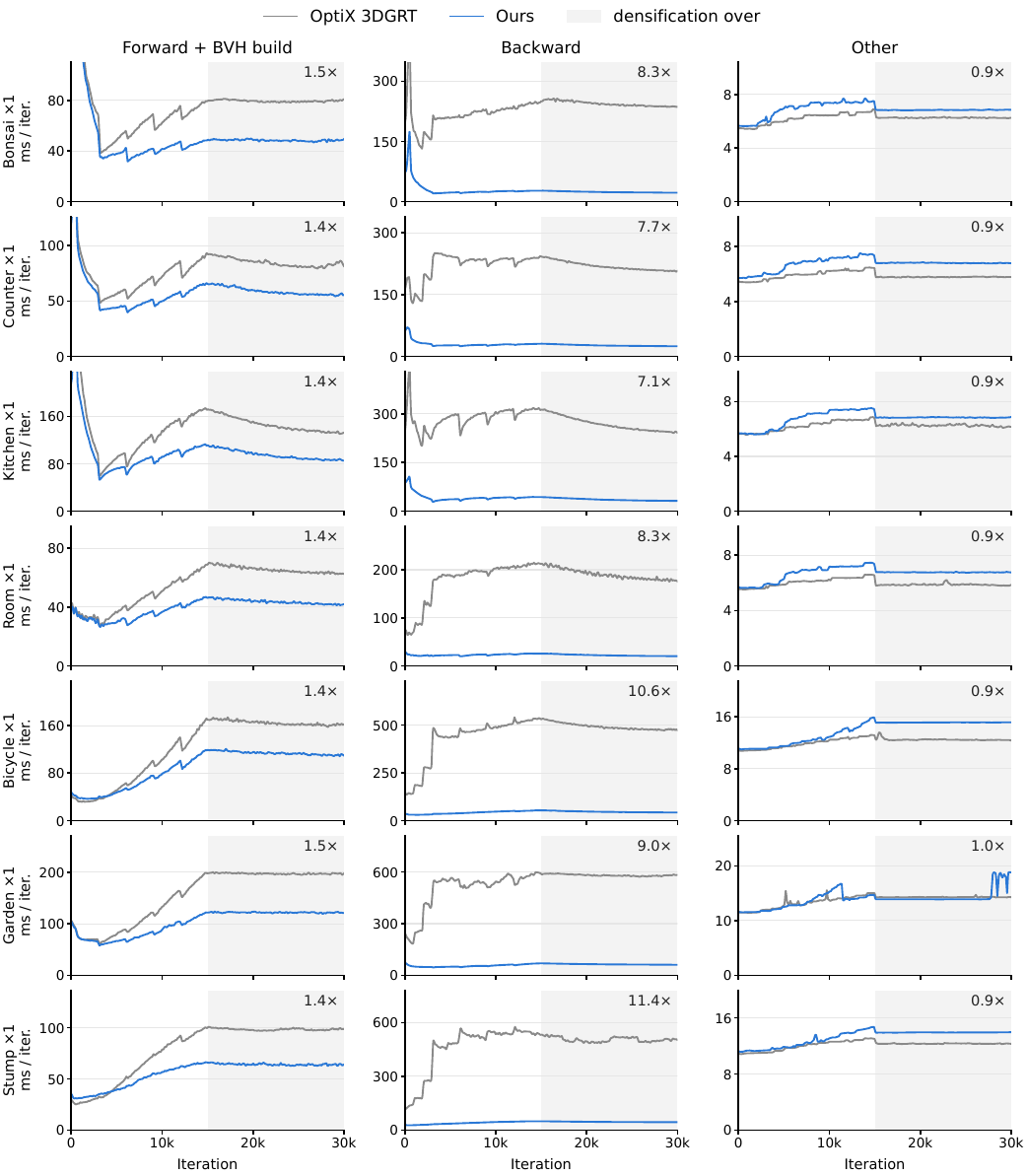}
    \caption{Per-phase time per training iteration, Mip-NeRF~360 at full resolution.}
    \label{fig:phases-full}
  \end{figure*}

    \begin{figure*}[t]
    \centering
    \includegraphics[width=\textwidth]{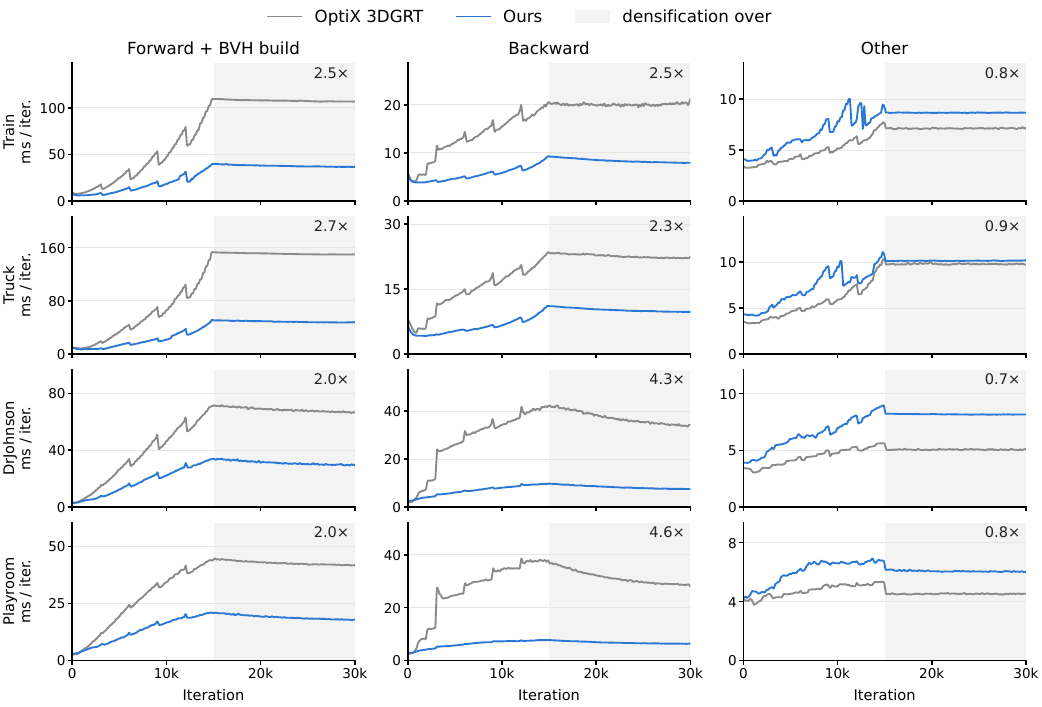}
    \caption{Per-phase time per training iteration, Tanks\&Temples and Deep Blending.}
    \label{fig:phases-other}
  \end{figure*}

\end{document}